\documentclass{article}

\usepackage{microtype}
\usepackage{graphicx}
\usepackage{subcaption}
\usepackage{booktabs}
\usepackage{hyperref}
\usepackage{amsmath}
\usepackage{amsthm}
\usepackage{bm}
\usepackage{amssymb}
\usepackage{mathtools}
\usepackage{amsthm}
\usepackage{multirow}
\usepackage{makecell}  
\usepackage{array}
\usepackage{xcolor}
\usepackage{colortbl}
\usepackage{float}    
\usepackage{makecell}
\usepackage{adjustbox}
\usepackage{pifont}

\usepackage[accepted]{icml2026}    
\usepackage[capitalize,noabbrev]{cleveref}

\theoremstyle{plain}

\theoremstyle{definition}

\theoremstyle{remark}

\icmltitlerunning{Bi-Level Sample Rebalancing with Pseudo-Label Diffusion for Point-Supervised Infrared Small-Target Detection}

\begin{document}

\twocolumn[
  \icmltitle{Diffuse to Detect: Bi-Level Sample Rebalancing with Pseudo-Label Diffusion for Point-Supervised Infrared Small-Target Detection}



  \icmlsetsymbol{equal}{*}

  \begin{icmlauthorlist}
    \icmlauthor{Zhu Liu}{DUT}
\icmlauthor{Yuanhang Yao}{DUT}
\icmlauthor{Ping Qian}{DUT}
\icmlauthor{Zihang Chen}{DUT}
\icmlauthor{Risheng Liu}{DUT}
  \end{icmlauthorlist}

\icmlaffiliation{DUT}{School of Software Technology, Dalian University of Technology, Dalian, China}

\icmlcorrespondingauthor{Risheng Liu}{rsliu@dlut.edu.cn}

  \icmlkeywords{Machine Learning, ICML}

  \vskip 0.3in
]



\printAffiliationsAndNotice{}  % no special notice (required even if empty)

\begin{abstract}
Point supervision has become a scalable solution to address dense annotation for infrared small target detection, but its performance is limited by two coupled bottlenecks: unstable pseudo-label evolution in cluttered, low-contrast infrared imagery and severe sample-distribution imbalance. In this paper, we present a more adaptive and stable framework to address these issues. Leveraging the intrinsic consistency between thermal radiation patterns and heat diffusion, we propose a physics-induced annotation strategy that expands single-point labels into reliable pseudo-masks. To further enhance supervision and alleviate sample imbalance, we develop a bi-level dual-update framework that jointly optimizes detector weights, sample weights, and diffusion parameters. A meta-classifier dynamically predicts sample-wise loss weights, while a differentiable diffusion module refines pseudo-labels with detection feedback, enabling adaptive interaction between training and hyperparameter optimization. Extensive experiments across multiple datasets demonstrate five-fold annotation acceleration, superior detection accuracy, and comparable performance with 30\% of the training data, validating the efficiency and practicality of our approach. Our code is available at \url{https://github.com/yuanhang-yao/diffuse-to-detect}.
\end{abstract}

\section{Introduction}
Infrared Small Target Detection (ISTD) is a long-standing and crucial task that aims to identify dim and tiny targets against cluttered backgrounds. As a component of infrared search-and-tracking systems, ISTD plays an indispensable role in monitoring~\cite{zhang2022isnet,li2022dense}, navigation~\cite{wang2025jtd}, early warning~\cite{xu2023multiscale}, and maritime resource management~\cite{zhang2022exploring}. However, due to the long-range imaging mechanism, targets are often shapeless and textureless, comprising only about 0.15\% of the total pixels, making them easily overlooked within complex and noisy backgrounds, which further complicates detection.

\begin{figure*}[thb]
	\centering
	\includegraphics[width=0.99\textwidth]{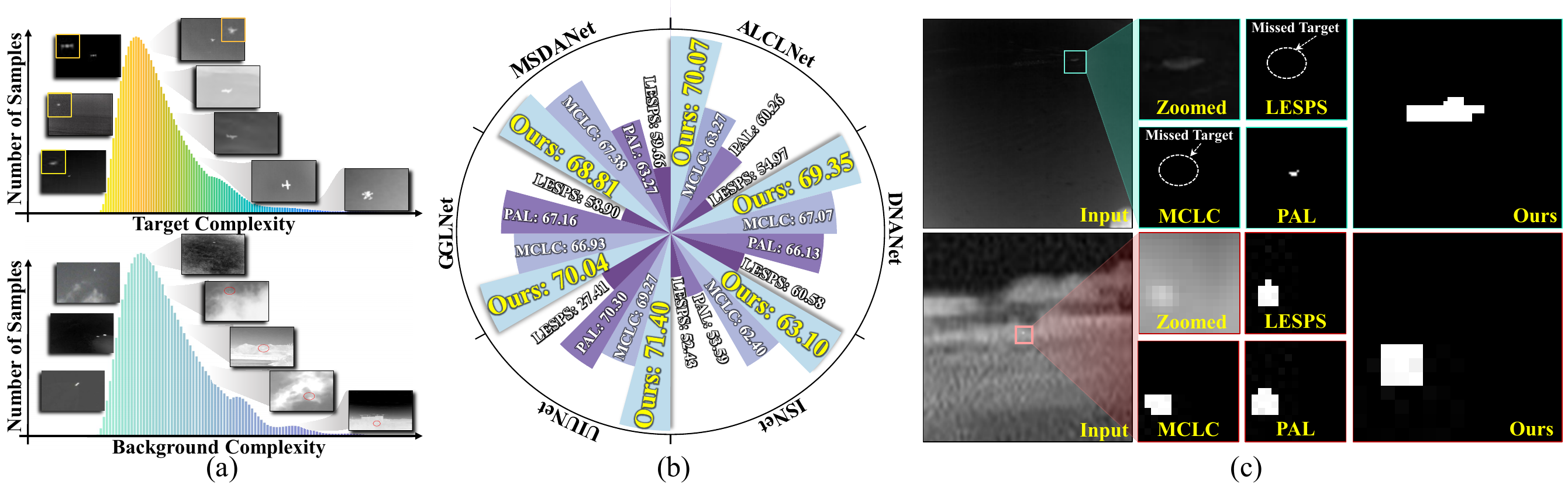}
	\caption{Motivation and efficiency overview. (a) Illustration of  challenges including sample imbalance and complexity. (b) Quantitative results showing clear improvements using only 30\% of the data. (c) Qualitative results validating our robustness in complex scenes.
	}
	\label{fig:first}
\end{figure*}

Recently, data-driven learning paradigms~\cite{dai2021asymmetric,dai2021attentional} have emerged as the mainstream solution, employing fully supervised pipelines to map infrared observations to pixel-level dense annotations. To capture discriminative features~\cite{jiang2025single}, existing methods design architectures guided by prior knowledge~\cite{zhang2025saist,Huang_2025_ICCV}, including densely nested structures~\cite{li2022dense}, vision transformers~\cite{chen2024tci}, or fine-tuning of large foundational models~\cite{zhang2024unleashing,zhang2024irsam}. Despite their promising performance, these approaches typically demand precise annotations and large-scale datasets covering diverse, high-stakes scenarios.
We argue that the development of learning-based ISTD faces two major obstacles. First, most approaches rely on pixel-level dense annotations~\cite{li2023monte}, which are costly to obtain and prone to errors due to the inherent characteristics of infrared small targets (\textit{e.g.}, low Signal-to-Noise Ratio (SNR) and indistinct boundaries). Second, the scarcity of high-quality ISTD datasets~\cite{liu2025deal} leads to severe long-tailed distributions, as illustrated in Fig.~\ref{fig:first} (a), \textit{e.g.}, only 200 and 800 training pairs in SIRST-v1 and IRSTD-1k. Such imbalance hinders sufficient training on rare but critical cases, thereby limiting generalization.

Existing studies~\cite{chen2013local,gao2013infrared} have explored single-point supervision as an alternative to alleviate the annotation burden. These approaches can be broadly divided into two promising directions. The first category employs online label updates. For example, LESPS~\cite{Ying_2023_CVPR} evolves pseudo-labels through intermediate indicators but suffers from excessive evolution and instability, while PAL~\cite{yu2025easy} integrates active learning with a coarse-to-fine training scheme to improve stability, yet still relies heavily on manual parameter tuning that limits flexibility. In contrast, offline update strategies generate pseudo-labels through dedicated algorithms~\cite{li2023monte,kou2024mcgc}, such as MCLC~\cite{li2023monte}, which constructs labels from repeated noisy perturbations based on Monte Carlo linear clustering. However, these methods remain vulnerable to false positives, where background noise is mistakenly identified as targets. These limitations highlight the need for a more stable and precise automatic pseudo-label generation strategy.

Furthermore, several attempts have been made to alleviate the extreme sample imbalance in ISTD. Static loss reweighting schemes, such as focal loss~\cite{xu2023multiscale}, truncated squeeze loss~\cite{zhang2025semi} and adversarial learning~\cite{wang2019miss} remain fixed throughout training. Curriculum-style methods, exemplified by progressive active learning, gradually introduce samples from easy to hard in a coarse-to-fine manner~\cite{yu2025easy}, yet still rely on manually crafted schedules and heuristic criteria. Despite their merits, these approaches lack adaptability to the evolving training dynamics.

To address the aforementioned challenges, we propose a more adaptive and stable ISTD framework under single-point supervision. Leveraging the natural consistency between radiation patterns of infrared targets and physical heat diffusion, we first introduce an offline physics-induced annotation strategy that treats each spot as a thermal source, with diffusion governing the spatial spread of energy. To further refine supervision and reduce errors in complex cases, we develop a bi-level dual-update framework from the perspective of hyperparameter optimization. In this framework, the inner-level optimization updates the detector weights by minimizing the training loss, while the outer-level optimization adaptively adjusts two types of hyperparameters: sample weights and diffusion parameters. Specifically, a meta-classifier predicts sample-wise loss weights conditioned on training dynamics, enabling adaptive rebalancing of sample contributions. In parallel, the diffusion process is parameterized as a differentiable module, whose parameters are updated by the outer-level objective to ensure pseudo-labels evolve consistently with validation feedback. Furthermore, we design a dynamic aggregated solution to coordinate the interaction between model training and hyperparameter optimization. Extensive experiments demonstrate that our approach accelerates annotation by five times, achieves superior performance across multiple benchmarks, and attains comparable results using only 30\% of the training data. The core contributions can be summarized as follows:
\begin{itemize}
	\item We propose a  bi-level dual-update optimization framework that jointly optimizes detector weights, sample weights, and annotation quality, achieving greater stability and adaptability than heuristic adjustments. 
    
	\item By integrating a meta-classifier for sample-wise loss weighting with a differentiable diffusion module, our framework simultaneously rebalances samples and refines pseudo-labels under detection feedback. 
    
	\item This work also introduces a physics-guided methodology for pseudo-mask generation, leveraging the intrinsic consistency between  radiation patterns and heat diffusion to achieve efficient single-point supervision.
	
	\item Extensive experiments across multiple ISTD benchmarks validate the efficiency (annotation acceleration), effectiveness (superior detection accuracy), and practicality (fewer training samples) of our approach.  
\end{itemize}

\begin{figure*}[!thb]
	\centering
	\setlength{\tabcolsep}{1pt} 
	\begin{tabular}{c}		
		\includegraphics[width=0.99\textwidth]{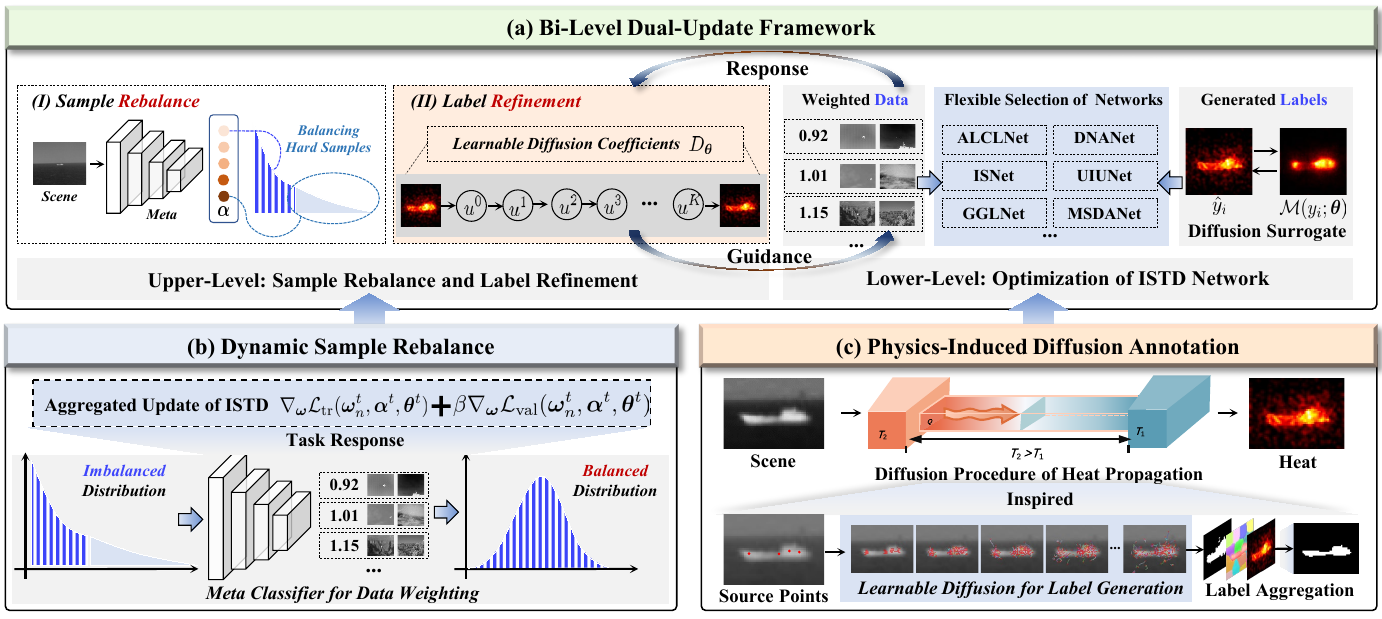}\\
	\end{tabular}
	\caption{\textcolor{black}{Overview of the proposed framework. (a) Bi-level dual-update framework performs joint sample rebalancing and label refinement. (b) Dynamic sample rebalancing is designed to weight training data. (c) Physics-induced diffusion annotation generates reliable pseudo-masks from single-point supervision in a learnable manner. }}
	\label{fig:workflow}
\end{figure*}

\section{Related Work}
\noindent\textbf{SIRST detection} has been widely studied for decades. Early research focused on model-driven traditional methods, including local contrast measurement methods~\cite{chen2013local,chen2014novel}, filtering-based methods~\cite{gao2013infrared}, and low-rank methods~\cite{he2015small,zhu2019infrared}. However, these methods require careful model design and fine-tuning of hyperparameters, resulting in poor adaptability and difficulty in dealing with complex and ever-changing real-world scenarios~\cite{liu2023paif}. The subsequent emergence of deep learning methods~\cite{zhao2022single,chen2023fluid}, especially fully supervised methods~\cite{lu2024sirst} based on customized architectures, has enabled models to learn the nonlinear mapping between input images~\cite{zhao2022single} and labels in a data-driven manner~\cite{liu2024infrared}. Various mechanisms including context aggregation~\cite{lu2025msca}, edge-guidance~\cite{li2024edge}, dense feature representation~\cite{li2022dense}, and spatial-frequency interaction, have been proposed~\cite{liu2025spatial}. Recently, several works have introduced granularity-aware modules and textual prompts into visual foundation models (\textit{e.g.,} SAM~\cite{kirillov2023segany}).
Unfortunately, these methods are limited by the scarcity of large-scale datasets, which limits generalization.

\noindent\textbf{Point-based supervision}
has been proposed for diverse visual perception tasks, such as object detection~\cite{Ying_2023_CVPR}, localization~\cite{li2022dense}, and segmentation. For instance, in oriented object detection,
Wholly-WOOD~\cite{yu2025wholly} studies unified weakly supervised oriented object detection with Point/HBox/RBox annotations, PointOBB~\cite{luo2024pointobb} focuses on recovering object scale and angle from point supervision for oriented detection, and Point2RBox-v2~\cite{yu2025point2rbox} improves point-supervised oriented detection by modeling spatial layout among instances.
At present, point-level segmentation methods require general targets with rich colors~\cite{gao2013infrared}, fine textures, and multiple annotated points, which pose challenges for the ISTD task. Ying \textit{et al.}~\cite{Ying_2023_CVPR} proposed the LESPS framework that implements single-point supervised SIRST detection. However, the LESPS framework has issues such as instability and excessive label evolution. Li \textit{et al.}~\cite{li2023monte} transformed fully supervised SIRST detection into a weakly supervised network with single-point supervision using the Monte Carlo clustering method, achieving remarkable performance while reducing annotation burden. Yu \textit{et al.}~\cite{yu2025easy} proposed the PAL framework, which enables the initial selection of simple samples and the generation of corresponding pseudo-labels, effectively alleviating the problem of excessive label evolution. However, it lacks adaptability to the constantly evolving training dynamics and still requires  adjustments. Thus, we aim to construct an efficient and adaptable framework.

\section{The Proposed Method}
\subsection{Bi-Level Dual-Update Framework}
In point-supervised ISTD, learning is fundamentally constrained by two coupled bottlenecks, including unstable pseudo-label evolution and severe sample-distribution imbalance.
Importantly, the foundational  issues amplify each other: imbalance makes the model more susceptible to overfitting to spurious background patterns in pseudo-labels, while noisy pseudo-labels further corrupt the estimation of sample importance, destabilizing the training dynamics.
To overcome these issues, we introduce a bi-level dual-update framework that jointly optimizes sample balance and supervision evolution based on the combination of offline and online updates, as shown in Fig.~\ref{fig:workflow} (a). 
Hyperparameter optimization  provides an effective tool for adaptively refining supervision according to model feedback, while the hierarchical separation enhances optimization stability and prevents error accumulation from noisy labels. Thus, we define the bi-level formulation~\cite{liu2024moreau,liu2024task} as:
\begin{equation}
	\begin{aligned}
		& \min_{\bm{\alpha},\bm{\theta}}\;\;
		\mathcal{L}_{\mathrm{val}} \big(\bm{\omega}^{*}(\bm{\alpha},\bm{\theta}),\bm{\alpha},\bm{\theta}\big) \label{eq:final_bi-level} \\ 	
		& \text{s.t.}\quad
		\bm{\omega}^{*}(\bm{\alpha},\bm{\theta})
		=
		\arg\min_{\bm{\omega}}
		\sum_{i}
		\alpha_i\,
		\mathcal{L}_{\mathrm{tr}} \big(\mathcal{N}(x_i;\bm{\omega}),\hat{y}_i\big),
	\end{aligned}
\end{equation}
where $\mathcal{N}$ denotes the detector with parameters $\bm{\omega}$, and $x_i$ and $\hat{y}_i(\bm{\theta})$ denote the input infrared image and its physics-induced pseudo-label, respectively. The variables $\bm{\alpha}$ and $\bm{\theta}$ correspond to two parallel hyperparameter branches in our framework.
Specifically, the inner-level problem updates the detector parameters $\bm{\omega}$ by minimizing the weighted training loss, where $\bm{\theta}$ controls the diffusion-based annotation process and $\bm{\alpha}$ denotes the sample-wise loss weights. The outer-level objective then refines both $\bm{\alpha}$ and $\bm{\theta}$ by minimizing the validation loss $\mathcal{L}_{\mathrm{val}}$, which reflects how well the detector generalizes under the current supervision (Fig.~\ref{fig:workflow} (b)). Note that, instead of leveraging real labels, we leverage the physics-induced pseudo-labels for the validation set. 
Compared with existing single-point supervised methods, the proposed bi-level dual-update framework establishes a dynamic closed-loop between model training and supervision generation, enabling continuous feedback to correct noisy labels and prevent overfitting.

\subsection{Physics-Induced Diffusion Annotation}
Infrared small targets are characterized by localized thermal emissions, small spatial size, and low signal-to-noise ratio (SNR). Most existing label-generation approaches often neglect the underlying physics of thermal propagation, failing to capture the spatial energy distribution. Physically, an infrared target acts as a local heat source, radiating energy that diffuses outward~\cite{liu2016learning,metzger2023guided}. This diffusion is intrinsically governed by the local thermal conductivity of the surrounding medium (\textit{e.g.,} texture and material), causing the energy to attenuate~\cite{borgnakke2020fundamentals}. This motivates our formulation of pseudo-label generation as a learnable thermal diffusion process. 
We model the pseudo-label $\hat{y}_i$ for the $i$-th target, identified by its single-point annotation $p_i$, as the quasi-steady thermal potential field $u(a,b)$ resulting from energy diffusion from the source $p_i$. The spatio-temporal evolution of this field, $u(a,b,t)$, is governed by a learnable anisotropic diffusion equation~\cite{bao2023heat,wang2023revelation}:
\begin{equation}\frac{\partial u(a,b,t)}{\partial t} =\nabla \cdot \big(D_{\bm{\theta}}(a,b)\nabla u(a,b,t)\big),\label{eq:pde_continuous}\end{equation}
with the initial condition $u(a,b,0)=\delta(a-p_i, b-p_i)$. Here, $D_{\bm{\theta}}(a,b)$ is the spatially varying diffusion tensor with learnable parameters $\bm{\theta}$. It governs the local thermal conductivity and modulates the diffusion rate according to regional texture and contrast: propagation accelerates in homogeneous areas (to fill missing regions) while decelerating near edges and high-gradient structures (to prevent boundary smearing). For computational tractability, we discretize Eq.~\eqref{eq:pde_continuous} into an iterative form:\begin{equation}u^{(k+1)} = (I-\tau L_{D_{\bm{\theta}}})u^{(k)},\quad u^{(0)}=\delta_{p_i},\label{eq:pde_discrete}\end{equation} where $u^{(k)}$ is the thermal potential after the $k$-th iteration, $\tau$ is the diffusion step size, and $L_{D_{\bm{\theta}}}$ is an image-adaptive Laplacian operator constructed from $D_{\bm{\theta}}(a,b)$. After $K$ iterations, $u^{(K)}$ represents the quasi-steady thermal field.

To enhance structural integrity and robustness, we aggregate this physics-based field with a data-driven spatial prior. Specifically, we introduce a superpixel segment $C_{p_i}(a,b)$, which is the binary mask of the superpixel region containing the annotation $p_i$. This prior provides a compact, data-driven boundary based on local image statistics. We then define the final pseudo-label $\hat{y}_i$ as a learnable aggregation of these two information sources:
\begin{equation}\hat{y}_i(\bm{\theta}) = \rho \cdot u^{(K)}(a,b) + (1-\rho) \cdot C_{p_i}(a,b),\label{eq:aggregation}\end{equation}
where $\rho \in [0, 1]$ is a learnable balance parameter. This fusion adaptively controls the trade-off between the physics-informed diffusion $u^{(K)}$ and the compact structural prior $C_{p_i}$. The aggregated field $\hat{y}_i(\bm{\theta})$ is subsequently normalized to $[0,1]$. The final pseudo-label $\hat{y}_i(\bm{\theta})$ thus preserves both physical continuity and structural precision, serving as a differentiable supervision signal for the bi-level optimization framework, as shown in Fig.~\ref{fig:workflow} (c).

\begin{table*}[thb]
	\centering
	\scriptsize
	\renewcommand{\arraystretch}{1.05}
    \caption{Performance of ISTD under different training strategies. All models are trained on SIRST3 and evaluated on SIRST3 and three individual datasets (SIRST-v1, NUDT-SIRST, IRSTD-1k). We report four metrics: IoU (\%), nIoU (\%), P$_{d}$ (\%), and F$_{a}$ ($10^{-6}$).}\label{tab:major1}
    \setlength{\tabcolsep}{1.2mm}
    \begin{tabular}{|c|c|cccc|cccc|cccc|cccc|}
        \hline                       
        && \multicolumn{4}{c|}{SIRST3} & \multicolumn{4}{c|}{SIRST-v1} & \multicolumn{4}{c|}{NUDT-SIRST} & \multicolumn{4}{c|}{IRSTD-1k} \\ \cline{3-18}
        \multirow{-2}{*}{Scheme} & \multirow{-2}{*}{Description} & IoU↑ & nIoU↑ & P$_{d}$↑ & F$_{a}$↓ & IoU↑ & nIoU↑ & P$_{d}$↑ & F$_{a}$↓ & IoU↑ & nIoU↑ & P$_{d}$↑ & F$_{a}$↓ & IoU↑ & nIoU↑ & P$_{d}$↑ & F$_{a}$↓  \\ \hline
                % -------- MLCLNet --------
        & Full  & 81.04 & 83.56 & 95.95 & 14.07 & 69.01 & 75.37 & 92.78 & 62.56 & 91.25 & 91.54 & 97.67 & 10.09 & 64.29 & 64.22 & 92.93 & 7.91 \\
        & LESPS & 35.02 & 44.81 & 69.50 & 28.22 & 44.93 & 52.68 & 77.95 & 33.79 & 38.71 & 46.31 & 70.16 & 27.16 & 26.75 & 41.49 & 71.04 & 22.75 \\
        & PAL   & 64.17 & \cellcolor[HTML]{D9E1F4}68.48 & \cellcolor[HTML]{D9E1F4}94.85 & \cellcolor[HTML]{D9E1F4}18.26 & 63.48 & \cellcolor[HTML]{D9E1F4}68.41 & \cellcolor[HTML]{D9E1F4}92.02 & \cellcolor[HTML]{D9E1F4}22.43 & 66.68 & \cellcolor[HTML]{D9E1F4}72.12 & \cellcolor[HTML]{D9E1F4}96.88 & \cellcolor[HTML]{D9E1F4}22.09 & \cellcolor[HTML]{FADADE}63.11 & \cellcolor[HTML]{D9E1F4}56.34 & \cellcolor[HTML]{D9E1F4}89.24 & 17.63 \\
        & MCLC  & \cellcolor[HTML]{D9E1F4}65.16 & 67.95 & \cellcolor[HTML]{FADADE}96.22 & 30.53 & \cellcolor[HTML]{D9E1F4}64.83 & 68.17 & 89.92 & 40.41 & \cellcolor[HTML]{D9E1F4}68.51 & 71.23 & 94.07 & 31.80 & 55.44 & 56.30 & 87.58 & \cellcolor[HTML]{D9E1F4}14.90 \\
        \cline{2-2}
        \multirow{-5}{*}{\makecell{MLCLNet\\\cite{yu2022infrared}}} & Ours & \cellcolor[HTML]{FADADE}69.95 & \cellcolor[HTML]{FADADE}72.61 & 94.57 & \cellcolor[HTML]{FADADE}11.12 & \cellcolor[HTML]{FADADE}65.70 & \cellcolor[HTML]{FADADE}69.39 & \cellcolor[HTML]{FADADE}92.59 & \cellcolor[HTML]{FADADE}7.90 & \cellcolor[HTML]{FADADE}78.68 & \cellcolor[HTML]{FADADE}85.70 & \cellcolor[HTML]{FADADE}97.67 & \cellcolor[HTML]{FADADE}12.79 & \cellcolor[HTML]{D9E1F4}60.19 & \cellcolor[HTML]{FADADE}56.94 & \cellcolor[HTML]{FADADE}90.12 & \cellcolor[HTML]{FADADE}11.20 \\ \hline
        
        % -------- ALCLNet --------
        & Full  & 82.12 & 82.57 & 96.15 & 13.36 & 75.65 & 75.91 & 96.20 & 24.90 & 88.78 & 89.26 & 97.67 & 10.23 & 65.87 & 64.31 & 91.25 & 12.13 \\
        & LESPS & 30.59 & 43.91 & 70.30 & \cellcolor[HTML]{D9E1F4}34.36 & 42.61 & 53.35 & 76.81 & 37.65 & 31.33 & 43.47 & 70.26 & \cellcolor[HTML]{D9E1F4}37.87 & 25.61 & 42.18 & 71.04 &27.27 \\
        & PAL   & 60.26 & 64.67 & \cellcolor[HTML]{D9E1F4}91.89 & 35.34 & 64.19 & 68.99 & 92.78 & \cellcolor[HTML]{D9E1F4}30.80 & \cellcolor[HTML]{D9E1F4}63.67 & \cellcolor[HTML]{D9E1F4}65.98 & \cellcolor[HTML]{D9E1F4}93.12 & 40.84 & 42.86 & 54.08 & 86.63 & \cellcolor[HTML]{D9E1F4}26.14 \\
        & MCLC  & \cellcolor[HTML]{D9E1F4}63.27 & \cellcolor[HTML]{D9E1F4}65.70 & \cellcolor[HTML]{D9E1F4}91.89 & 47.19 & \cellcolor[HTML]{D9E1F4}68.09 & \cellcolor[HTML]{D9E1F4}72.41 & \cellcolor[HTML]{D9E1F4}93.54 & 51.86 & 63.01 & 65.58 & 91.22 & 49.25 & \cellcolor[HTML]{D9E1F4}58.15 & \cellcolor[HTML]{D9E1F4}55.94 & \cellcolor[HTML]{D9E1F4}88.26 & 34.47 \\
        \cline{2-2}
        \multirow{-5}{*}{\makecell{ALCLNet\\\cite{yu2022pay}}} & Ours & \cellcolor[HTML]{FADADE}73.28 & \cellcolor[HTML]{FADADE}73.30 & \cellcolor[HTML]{FADADE}95.44 & \cellcolor[HTML]{FADADE}16.35 & \cellcolor[HTML]{FADADE}74.34 & \cellcolor[HTML]{FADADE}72.92 & \cellcolor[HTML]{FADADE}98.15 & \cellcolor[HTML]{FADADE}13.82 & \cellcolor[HTML]{FADADE}79.94 & \cellcolor[HTML]{FADADE}85.50 & \cellcolor[HTML]{FADADE}97.67 & \cellcolor[HTML]{FADADE}10.04 & \cellcolor[HTML]{FADADE}63.96 & \cellcolor[HTML]{FADADE}57.66 & \cellcolor[HTML]{FADADE}92.18 & \cellcolor[HTML]{FADADE}24.29 \\ \hline
        
        % -------- DNANet --------
        & Full  & 85.49 & 86.22 & 96.88 & 8.14 & 77.95 & 80.35 & 96.58 & 22.23 & 93.90 & 94.06 & 98.84 & 3.06 & 65.21 & 66.64 & 91.25 & 10.21 \\
        & LESPS & 36.98 & 47.59 & 86.84 & 28.97 & 44.39 & 52.49 & 84.79 & 30.10 & 37.27 & 46.40 & 90.58 & 26.04 & 31.97 & 47.92 & 76.43 &27.80 \\
        & PAL   & 66.13 & \cellcolor[HTML]{D9E1F4}70.37 & \cellcolor[HTML]{D9E1F4}94.76 & \cellcolor[HTML]{D9E1F4}21.38 & 67.28 & \cellcolor[HTML]{D9E1F4}70.91 & 95.06 & \cellcolor[HTML]{D9E1F4}22.09 & \cellcolor[HTML]{D9E1F4}71.23 & \cellcolor[HTML]{D9E1F4}73.53 & \cellcolor[HTML]{D9E1F4}96.88 & \cellcolor[HTML]{D9E1F4}22.24 & 48.89 & \cellcolor[HTML]{D9E1F4}57.38 & 89.27 & 23.85 \\
        & MCLC  & \cellcolor[HTML]{D9E1F4}67.07 & 69.29 & 94.22 & 33.03 & \cellcolor[HTML]{D9E1F4}68.08 & 69.08 & \cellcolor[HTML]{D9E1F4}95.96 & 37.32 & 70.22 & 73.03 & 94.71 & 35.46 & \cellcolor[HTML]{D9E1F4}56.92 & 56.66 & \cellcolor[HTML]{D9E1F4}90.57 &  \cellcolor[HTML]{D9E1F4}23.55 \\
        \cline{2-2}
        \multirow{-5}{*}{\makecell{DNANet\\\cite{li2022dense}}} & Ours & \cellcolor[HTML]{FADADE}71.08 & \cellcolor[HTML]{FADADE}74.65 & \cellcolor[HTML]{FADADE}95.24 & \cellcolor[HTML]{FADADE}15.16 & \cellcolor[HTML]{FADADE}68.27 & \cellcolor[HTML]{FADADE}73.49 & \cellcolor[HTML]{FADADE}96.30 & \cellcolor[HTML]{FADADE}14.90 & \cellcolor[HTML]{FADADE}79.99 & \cellcolor[HTML]{FADADE}87.01 & \cellcolor[HTML]{FADADE}97.67 & \cellcolor[HTML]{FADADE}13.65 & \cellcolor[HTML]{FADADE}61.38 & \cellcolor[HTML]{FADADE}58.07 & \cellcolor[HTML]{FADADE}90.82 & \cellcolor[HTML]{FADADE}17.00 \\ \hline

        & Full & 74.38 & 76.34 & 93.29 & 28.10 & 70.47 & 72.89 & 92.78 & 41.23 & 80.19 & 81.73 & 94.18 & 22.41 & 61.46 & 60.52 & 89.23 & 32.13 \\
        & LESPS & 32.22 & 44.20 & 77.94 & 36.98 & 40.71 & 51.00 & 82.13 & 38.59 & 35.46 & 45.63 & 79.15 & 35.02 & 24.02 & 38.40 & 71.72 & 36.74 \\
        & PAL & 53.59 & 58.42 & 86.11 & \cellcolor[HTML]{D9E1F4}35.57 & 53.01 & 60.53 & 88.21 & 32.38 & 57.69 & 62.15 & 88.68 & \cellcolor[HTML]{D9E1F4}34.75 & 31.92 & 45.23 & 77.78 & 39.89 \\
        & MCLC & \cellcolor[HTML]{D9E1F4}62.40 & \cellcolor[HTML]{D9E1F4}64.07 & \cellcolor[HTML]{D9E1F4}90.42 & 45.53 & \cellcolor[HTML]{D9E1F4}65.12 &\cellcolor[HTML]{D9E1F4}68.04 & \cellcolor[HTML]{D9E1F4}94.68 & \cellcolor[HTML]{D9E1F4}27.71 & \cellcolor[HTML]{D9E1F4}65.13 & \cellcolor[HTML]{D9E1F4}66.88 &\cellcolor[HTML]{D9E1F4}92.76 & 55.59 & \cellcolor[HTML]{D9E1F4}52.65 & \cellcolor[HTML]{D9E1F4}52.35 & \cellcolor[HTML]{D9E1F4}85.58 &\cellcolor[HTML]{D9E1F4}32.57 \\
        \cline{2-2} 
        \multirow{-5}{*}{\makecell{ISNet\\\cite{zhang2022isnet}}} & Ours &\cellcolor[HTML]{FADADE}65.83 &\cellcolor[HTML]{FADADE}69.39 &\cellcolor[HTML]{FADADE}91.02 & \cellcolor[HTML]{FADADE}19.29 & \cellcolor[HTML]{FADADE}66.82 &\cellcolor[HTML]{FADADE}71.14 & \cellcolor[HTML]{FADADE}96.30 & \cellcolor[HTML]{FADADE}23.21 & \cellcolor[HTML]{FADADE}72.60 & \cellcolor[HTML]{FADADE}79.91 & \cellcolor[HTML]{FADADE}93.02 & \cellcolor[HTML]{FADADE}9.24 &\cellcolor[HTML]{FADADE}54.81 &\cellcolor[HTML]{FADADE}53.53 & \cellcolor[HTML]{FADADE}86.39 & \cellcolor[HTML]{FADADE}30.59 \\ \hline
        & Full & 85.21 & 85.73 & 97.01 & 11.86 & 76.74 & 78.53 & 95.82 & 21.82 & 93.57 & 93.75 & 98.62 & 4.69 & 68.22 & 66.57 & 93.27 & 20.95 \\
        & LESPS & 39.38 & 50.20 & 75.81 & 33.43 & 53.06 & 60.58 & 82.89 & \cellcolor[HTML]{D9E1F4}29.76 & 40.17 & 49.04 & 75.24 & 30.32 & 32.37 & 47.56 & 76.09 & 36.25 \\
        & PAL & \cellcolor[HTML]{D9E1F4}70.30 & \cellcolor[HTML]{D9E1F4}71.95 & 94.48 & \cellcolor[HTML]{D9E1F4}25.29 & 72.15 &\cellcolor[HTML]{D9E1F4}74.15 & 95.44 & 34.64 &\cellcolor[HTML]{D9E1F4}71.95 & \cellcolor[HTML]{D9E1F4}73.71 & \cellcolor[HTML]{D9E1F4}97.31 &\cellcolor[HTML]{D9E1F4}20.87 &\cellcolor[HTML]{D9E1F4}65.82 & 62.21 &\cellcolor[HTML]{D9E1F4}91.95 &\cellcolor[HTML]{D9E1F4}31.35 \\
        & MCLC & 69.27 & 71.07 & \cellcolor[HTML]{D9E1F4}94.81 & 37.60 & \cellcolor[HTML]{D9E1F4}72.45 & 73.76 & \cellcolor[HTML]{D9E1F4}96.20 & 42.26 & 69.81 & 73.35 & 96.08 & 36.97 & 65.15 & \cellcolor[HTML]{D9E1F4}63.79 & 91.93 & 36.31 \\
        \cline{2-2}
        \multirow{-5}{*}{\makecell{UIUNet\\\cite{wu2022uiu}}} & Ours &\cellcolor[HTML]{FADADE}73.65 & \cellcolor[HTML]{FADADE}75.60 &\cellcolor[HTML]{FADADE}96.31 & \cellcolor[HTML]{FADADE}15.40 & \cellcolor[HTML]{FADADE}73.00 & \cellcolor[HTML]{FADADE}74.59 & \cellcolor[HTML]{FADADE}100.00 &\cellcolor[HTML]{FADADE}11.49 & \cellcolor[HTML]{FADADE}79.33 &\cellcolor[HTML]{FADADE}84.22 & \cellcolor[HTML]{FADADE}97.67 & \cellcolor[HTML]{FADADE}15.95 & \cellcolor[HTML]{FADADE}65.86 & \cellcolor[HTML]{FADADE}64.75 & \cellcolor[HTML]{FADADE}92.16 & \cellcolor[HTML]{FADADE}16.93 \\ \hline
        
        & Full & 84.08 & 85.38 & 97.61 & 8.58 & 78.82 & 79.45 & 97.34 & 25.93 & 92.33 & 92.86 & 99.26 & 3.65 & 64.52 & 65.54 & 91.58 & 8.27 \\
        & LESPS & 40.89 & 50.95 & 74.42 & 33.34 & 50.07 & 59.07 & 79.85 & 33.94 & 43.61 & 51.04 & 76.19 & 37.03 & 32.26 & 46.75 & 71.04 & 28.67 \\
        & PAL & \cellcolor[HTML]{D9E1F4}67.16 & \cellcolor[HTML]{D9E1F4}70.42 & \cellcolor[HTML]{D9E1F4}94.95 & 27.19 & 68.81 & 71.78 & 96.20 & 27.51 & \cellcolor[HTML]{D9E1F4}71.07 & \cellcolor[HTML]{D9E1F4}72.71 & \cellcolor[HTML]{D9E1F4}97.46 & 26.93 & 53.08 & \cellcolor[HTML]{D9E1F4}59.54 & \cellcolor[HTML]{D9E1F4}89.23 & \cellcolor[HTML]{D9E1F4}27.19 \\
        & MCLC & 66.93 & 68.64 & 93.15 & \cellcolor[HTML]{D9E1F4}26.02 & \cellcolor[HTML]{D9E1F4}70.20 & \cellcolor[HTML]{D9E1F4}71.94 & \cellcolor[HTML]{D9E1F4}97.72 & \cellcolor[HTML]{D9E1F4}21.40 & 68.24 & 70.26 & 93.23 & \cellcolor[HTML]{D9E1F4}26.34 & \cellcolor[HTML]{D9E1F4}56.82 & 58.87 & 88.92 & 29.98 \\
        \cline{2-2}
        \multirow{-5}{*}{\makecell{GGLNet\\\cite{zhao2023gradient}}} & Ours &\cellcolor[HTML]{FADADE}71.73 & \cellcolor[HTML]{FADADE}75.35 & \cellcolor[HTML]{FADADE}95.04 & \cellcolor[HTML]{FADADE}16.23 & \cellcolor[HTML]{FADADE}72.55 &\cellcolor[HTML]{FADADE}73.37 & \cellcolor[HTML]{FADADE}98.15 & \cellcolor[HTML]{FADADE}11.13 &\cellcolor[HTML]{FADADE}82.20 & \cellcolor[HTML]{FADADE}88.33 &\cellcolor[HTML]{FADADE}98.26 & \cellcolor[HTML]{FADADE}14.63 &\cellcolor[HTML]{FADADE}58.24 & \cellcolor[HTML]{FADADE}59.91 & \cellcolor[HTML]{FADADE}89.80 & \cellcolor[HTML]{FADADE}20.48 \\ \hline
        
        & Full & 86.16 & 86.13 & 97.01 & 11.04 & 76.40 & 78.03 & 94.30 & 25.79 & 94.01 & 94.18 & 98.73 & 1.47 & 71.64 & 68.09 & 93.94 & 28.26 \\
        & LESPS & 39.03 & 49.84 & 82.86 & 25.58 & 50.27 & 58.15 & 85.17 & 22.43 & 39.60 & 48.67 & 84.23 & 24.82 & 33.11 & 47.70 & 78.11 & 29.42 \\
        & PAL & 63.27 & 68.68 & \cellcolor[HTML]{D9E1F4}94.15 & \cellcolor[HTML]{D9E1F4}23.29 & 67.76 & 71.34 & 94.68 & 16.81 & 67.48 & 70.61 & \cellcolor[HTML]{D9E1F4}96.83 & \cellcolor[HTML]{D9E1F4}23.76 & 47.80 & 58.02 & 89.23 & 27.54 \\
        & MCLC &\cellcolor[HTML]{D9E1F4}67.38 & \cellcolor[HTML]{D9E1F4}69.88 & 93.82 & 37.35 & \cellcolor[HTML]{D9E1F4}70.61 & \cellcolor[HTML]{D9E1F4}72.31 & \cellcolor[HTML]{D9E1F4}96.96 &\cellcolor[HTML]{D9E1F4}14.13 &\cellcolor[HTML]{D9E1F4}68.94 &\cellcolor[HTML]{D9E1F4}71.67 & 94.81 & 47.68 & \cellcolor[HTML]{D9E1F4}58.84 &\cellcolor[HTML]{D9E1F4}58.71 & \cellcolor[HTML]{D9E1F4}89.93 & \cellcolor[HTML]{D9E1F4}27.33 \\
        \cline{2-2} 
        \multirow{-5}{*}{\makecell{MSDANet\\\cite{zhao2024multiscaledirectionawarenetworkinfrared}}} & Ours &\cellcolor[HTML]{FADADE}71.41 &\cellcolor[HTML]{FADADE}73.16 & \cellcolor[HTML]{FADADE}94.37 & \cellcolor[HTML]{FADADE}12.71 &\cellcolor[HTML]{FADADE}71.99 & \cellcolor[HTML]{FADADE}72.51 & \cellcolor[HTML]{FADADE}97.22 & \cellcolor[HTML]{FADADE}7.54 &\cellcolor[HTML]{FADADE}79.25 & \cellcolor[HTML]{FADADE}85.38 & \cellcolor[HTML]{FADADE}97.09 & \cellcolor[HTML]{FADADE}5.85 & \cellcolor[HTML]{FADADE}60.61 & \cellcolor[HTML]{FADADE}59.38 & \cellcolor[HTML]{FADADE}90.82 &\cellcolor[HTML]{FADADE}23.76 \\ \hline
    \end{tabular}
\end{table*}

\begin{figure*}[thbp]
    \centering
    \includegraphics[width=0.99\textwidth, height=0.38\textwidth]{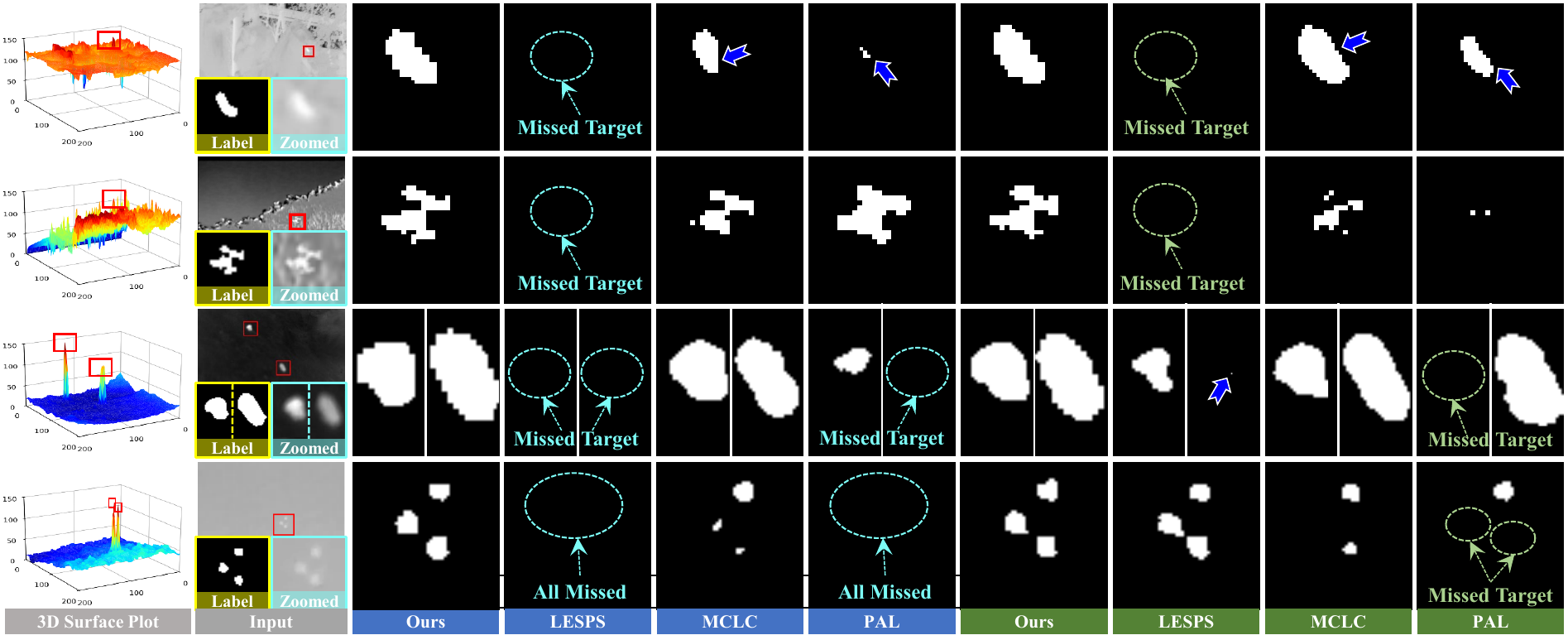}
    \caption{Qualitative comparison on the SIRST3 dataset. Columns from left to right: 3D surfaces of inputs, input images with labels and local magnification, Ours, LESPS, MCLC, and PAL predictions (the blue part on ALCLNet, the green part on DNANet). }
    \label{fig:radiation1}
\end{figure*}

\subsection{Dynamic Aggregated Solution}
In practice, solving the hyperparameter optimization in Eq.~\eqref{eq:final_bi-level} is computationally demanding, as it typically requires high-order unrolled gradients~\cite{liu2025augmenting}
or implicit differentiation~\cite{chen2022gradient}. Although alternating updates can reduce computational burden, they often lead to unstable convergence due to the mismatched update frequencies between the inner and outer levels. 
To overcome these limitations, we introduce a first-order dynamic aggregated solution~\cite{liu2023bi} that enables efficient and coherent interaction between model parameters and hyperparameters within a unified optimization. 

Inspired by recent work on single-loop bi-level optimization~\cite{jiang2025beyond},
we dynamically combine these two objectives to update the model parameters in the inner loop to obtain $\bm{\omega}^{t+1}$: $
    \bm{\omega}^{t}_{n+1} = \bm{\omega}^{t}_{n} - \Big( \nabla_{\bm{\omega}}\mathcal{L}_{\mathrm{tr}}(\bm{\omega}^{t}_{n},\bm{\alpha}^{t},\bm{\theta}^{t}) + \beta\,\nabla_{\bm{\omega}}\mathcal{L}_{\mathrm{val}}(\bm{\omega}^{t}_{n},\bm{\alpha}^{t},\bm{\theta}^{t}) \Big)$,
where $\beta$ controls how strongly the outer-level objective influences the inner-level parameter updates, achieving a balance between leveraging validation feedback and maintaining training stability.
This dynamic aggregation allows the model parameters to evolve under both 
task-specific supervision and meta-level  feedback.
Subsequently, the hyperparameters $\bm{\alpha}$ are updated in a one-step gradient manner 
based on the latest model parameters $\bm{\omega}^{t+1}$:
$\bm{\alpha}^{t+1}
=
\bm{\alpha}^{t}
-
\nabla_{\bm{\alpha}}\mathcal{L}_{\mathrm{val}}(\bm{\omega}^{t+1},\bm{\alpha}^{t},\bm{\theta}^{t}).$
This update strategy ensures that the learned loss weights are adaptively 
aligned with the outer-level objectives.
Finally, we refine the diffusion parameters $\bm{\theta}$ by aligning the differentiable diffusion surrogate $\mathcal{M}$ with the physics-induced pseudo-labels produced by the offline annotator. 
Given the detector prediction $y_i = \mathcal{N}(x_i;\bm{\omega})$, the surrogate diffusion module outputs a soft mask $\mathcal{M}(y_i;\bm{\theta})$, while $\hat{y}_i$ denotes the corresponding pseudo-label generated by the basic physics-induced diffusion process.
We update $\bm{\theta}$ with a single gradient step:
$
\bm{\theta}^{t+1}
=
\bm{\theta}^{t}
-
\nabla_{\bm{\theta}}
\Big(
\mathcal{L}_\mathrm{val}\big(
\mathcal{M}(y_i;\bm{\theta}),\,\hat{y}_i
\big)
+
\mathcal{L}_\mathrm{val}\big(
\mathcal{M}(y_i;\bm{\theta}),\,y_i
\big)
\Big).
$
Here, the first term encourages $\mathcal{M}$ to reproduce the physics-induced pseudo-labels, while the second term regularizes the update by keeping $\mathcal{M}(y_i;\bm{\theta})$ consistent with the current  response, mitigating overfitting to potentially noisy supervision.
This design ensures that the outer objective provides a stable signal for generalization-oriented weighting and hyperparameter calibration, while preventing the validation loss from collapsing into self-consistency with the current predictions.

\begin{table*}[!thb]
    \centering
    \scriptsize
    \renewcommand{\arraystretch}{1.05}
    \caption{Performance of ISTD under different training strategies. Models are trained on three individual datasets (SIRST-v1, NUDT-SIRST, IRSTD-1k). We report four metrics: IoU (\%), nIoU (\%), P$_{d}$ (\%), and F$_{a}$ ($10^{-6}$).} \label{tab:major2}
    \setlength{\tabcolsep}{2.5mm}{
    \begin{tabular}{|c|c|cccc|cccc|cccc|}
        \hline
        &                               & \multicolumn{4}{c|}{SIRST-v1}                                                                                             & \multicolumn{4}{c|}{NUDT-SIRST}                                                                                               & \multicolumn{4}{c|}{IRSTD-1k}                                                                                                 \\ \cline{3-14} 
        \multirow{-2}{*}{Methods}  & \multirow{-2}{*}{Strategy} & IoU↑                          & nIoU↑                         & P$_{d}$↑                            & F$_{a}$↓                           & IoU↑                          & nIoU↑                         & P$_{d}$↑                           & F$_{a}$↓                           & IoU↑                          & nIoU↑                         & P$_{d}$↑                           & F$_{a}$↓                           \\ \hline
        & Full                          & 72.57                         & 72.66                         & 94.68                          & 30.39                         & 89.79                         & 89.95                         & 98.84                         & 6.69                          & 65.23                         & 64.25                         & 89.90                         & 9.19                          \\
        & LESPS                        & 39.85 & 43.16 & 73.76 & 35.37 & 36.76 & 44.90 & 73.02 & 39.02 & 33.73 & 47.75 & 80.81 &36.21 \\
        & PAL                           & 60.49                         & 62.48                         & 86.69                          & 43.63                         & 67.40                         & 68.89                         & \cellcolor[HTML]{D9E1F4}96.61 & \cellcolor[HTML]{D9E1F4}23.00 & 57.48                         & 55.40                         & 87.54                         & \cellcolor[HTML]{D9E1F4}27.86 \\
        & MCLC                          & \cellcolor[HTML]{D9E1F4}71.80 & \cellcolor[HTML]{D9E1F4}71.71 & \cellcolor[HTML]{D9E1F4}95.06  & \cellcolor[HTML]{D9E1F4}33.07 & \cellcolor[HTML]{D9E1F4}68.89 & \cellcolor[HTML]{D9E1F4}70.84 & 95.45                         & 39.94                         & \cellcolor[HTML]{D9E1F4}63.11 & \cellcolor[HTML]{D9E1F4}59.66 & \cellcolor[HTML]{D9E1F4}91.58 & 29.80                         \\ \cline{2-14} 
        \multirow{-5}{*}{\makecell{ALCLNet\\\cite{yu2022pay}}} & Ours                          & \cellcolor[HTML]{FADADE}75.62 & \cellcolor[HTML]{FADADE}71.93 & \cellcolor[HTML]{FADADE}98.15  & \cellcolor[HTML]{FADADE}2.51  & \cellcolor[HTML]{FADADE}79.88 & \cellcolor[HTML]{FADADE}85.50 & \cellcolor[HTML]{FADADE}96.80 & \cellcolor[HTML]{FADADE}17.27 & \cellcolor[HTML]{FADADE}64.72 & \cellcolor[HTML]{FADADE}60.35 & \cellcolor[HTML]{FADADE}92.52 & \cellcolor[HTML]{FADADE}18.07 \\ \hline
        
        & Full                          & 77.55                         & 77.76                         & 94.30                          & 11.66                         & 95.35                         & 95.43                         & 99.26                         & 2.00                          & 67.74                         & 63.66                         & 88.55                         & 17.56                         \\
        & LESPS                         & 54.39 & 61.13 & 89.35 & 30.18 & 36.98 & 46.46 & 66.14 & 31.97 & 30.59 & 47.47 & 72.39 &29.60 \\
        & PAL                           & 64.83                         & 65.94                         & 92.78                          & 30.87                         & 70.70                         & 71.34                         & \cellcolor[HTML]{D9E1F4}97.78 & \cellcolor[HTML]{D9E1F4}24.57 & \cellcolor[HTML]{D9E1F4}62.29 & 58.46                         & 87.88                         & \cellcolor[HTML]{D9E1F4}20.65 \\
        & MCLC                          & \cellcolor[HTML]{D9E1F4}69.61 & \cellcolor[HTML]{D9E1F4}72.21 & \cellcolor[HTML]{D9E1F4}96.96  & \cellcolor[HTML]{D9E1F4}24.08 & \cellcolor[HTML]{D9E1F4}70.86 & \cellcolor[HTML]{D9E1F4}72.82 & 96.19                         & 25.00                         & 61.79                         & \cellcolor[HTML]{D9E1F4}64.32 & \cellcolor[HTML]{D9E1F4}90.91 & 32.51                         \\ \cline{2-14} 
        \multirow{-5}{*}{\makecell{DNANet\\\cite{li2022dense}}}  & Ours                          & \cellcolor[HTML]{FADADE}75.49 & \cellcolor[HTML]{FADADE}72.94 & \cellcolor[HTML]{FADADE}99.07  & \cellcolor[HTML]{FADADE}8.08  & \cellcolor[HTML]{FADADE}79.32 & \cellcolor[HTML]{FADADE}87.20 & \cellcolor[HTML]{FADADE}98.26 & \cellcolor[HTML]{FADADE}10.21 & \cellcolor[HTML]{FADADE}63.64 & \cellcolor[HTML]{FADADE}64.67 & \cellcolor[HTML]{FADADE}92.59 & \cellcolor[HTML]{FADADE}18.67 \\ \hline

        & Full                          & 80.03                         & 78.99                         & 94.68                          & 9.12                          & 95.38                         & 95.22                         & 99.05                         & 1.38                          & 70.36                         & 64.09                         & 93.60                         & 18.24                         \\
        & LESPS                         & 46.82 & 55.18 &78.71 & 26.40 & 39.63 & 46.56 & 68.04 & 20.04 & 39.90 & 49.67 & 80.47 & 20.70 \\
        & PAL                           & 63.98                         & 62.39                         & \cellcolor[HTML]{D9E1F4}92.02  & 28.95                         & \cellcolor[HTML]{D9E1F4}70.81 & \cellcolor[HTML]{D9E1F4}72.43 & \cellcolor[HTML]{D9E1F4}98.20 & \cellcolor[HTML]{D9E1F4}13.42 & 57.50                         & 56.98                         & \cellcolor[HTML]{D9E1F4}89.90 & \cellcolor[HTML]{D9E1F4}11.86 \\
        & MCLC                          & \cellcolor[HTML]{D9E1F4}71.01 & \cellcolor[HTML]{D9E1F4}68.87 & 90.11                          & \cellcolor[HTML]{D9E1F4}22.91 & 70.20                         & 72.22                         & 94.71                         & 21.46                         & \cellcolor[HTML]{D9E1F4}66.19 & \cellcolor[HTML]{D9E1F4}62.31 & 88.89                         & 39.30                         \\ \cline{2-14} 
        \multirow{-5}{*}{\makecell{UIUNet\\\cite{wu2022uiu}}}  & Ours                          & \cellcolor[HTML]{FADADE}71.85 & \cellcolor[HTML]{FADADE}69.55 & \cellcolor[HTML]{FADADE}99.07  & \cellcolor[HTML]{FADADE}11.31 & \cellcolor[HTML]{FADADE}79.84 & \cellcolor[HTML]{FADADE}85.19 & \cellcolor[HTML]{FADADE}98.55 & \cellcolor[HTML]{FADADE}6.77  & \cellcolor[HTML]{FADADE}68.19 & \cellcolor[HTML]{FADADE}62.86 & \cellcolor[HTML]{FADADE}90.14 & \cellcolor[HTML]{FADADE}6.07  \\ \hline
        & Full                          & 75.82                         & 76.49                         & 95.06                          & 38.49                         & 95.65                         & 95.48                         & 99.47                         & 1.38                          & 68.08                         & 64.29                         & 91.92                         & 24.05                         \\
        & LESPS                         & 48.69 & 55.86 & 82.89 & 32.12 & 36.34 & 45.45 & 73.76 & 19.40 & 38.73 & 49.79 & 81.48 & 33.55 \\
        & PAL                           & 63.27                         & 66.92                         & \cellcolor[HTML]{D9E1F4}93.92  & \cellcolor[HTML]{D9E1F4}29.64 & 59.46                         & 63.76                         & 93.54                         & \cellcolor[HTML]{D9E1F4}16.13 & 59.52                         & 58.16                         & \cellcolor[HTML]{D9E1F4}89.90 & \cellcolor[HTML]{D9E1F4}26.94 \\
        & MCLC                          & \cellcolor[HTML]{D9E1F4}71.23 & \cellcolor[HTML]{D9E1F4}72.35 & 92.02                          & 31.83                         & \cellcolor[HTML]{D9E1F4}69.67 & \cellcolor[HTML]{D9E1F4}71.61 & \cellcolor[HTML]{D9E1F4}94.92 & 26.52                         & \cellcolor[HTML]{D9E1F4}62.83 & \cellcolor[HTML]{D9E1F4}61.77 & 87.54                         & 32.32                         \\ \cline{2-14} 
        \multirow{-5}{*}{\makecell{MSDANet\\\cite{zhao2024multiscaledirectionawarenetworkinfrared}}} & Ours                          & \cellcolor[HTML]{FADADE}76.33 & \cellcolor[HTML]{FADADE}73.79 & \cellcolor[HTML]{FADADE}100.00 & \cellcolor[HTML]{FADADE}3.77  & \cellcolor[HTML]{FADADE}79.08 & \cellcolor[HTML]{FADADE}86.05 & \cellcolor[HTML]{FADADE}98.84 & \cellcolor[HTML]{FADADE}13.19 & \cellcolor[HTML]{FADADE}63.31 & \cellcolor[HTML]{FADADE}62.94 & \cellcolor[HTML]{FADADE}90.12 & \cellcolor[HTML]{FADADE}25.98 \\ \hline
    \end{tabular}}
\end{table*}

\section{Experiments}
\subsection{Implementation Configurations}
\noindent\textbf{Datasets.} We evaluate on four representative  datasets: SIRST3~\cite{Ying_2023_CVPR}, SIRST-v1~\cite{dai2021asymmetric}, NUDT-SIRST~\cite{li2022dense}, and IRSTD-1k~\cite{zhang2022isnet}, which contain 2,755, 427, 1,327, and 1,001 samples, respectively.
Since there is no common practice of using a validation set on these benchmarks, we adopt a 6:2:2 partition for each of SIRST-v1, NUDT-SIRST, and IRSTD-1k, following the official splitting protocols~\cite{dai2021asymmetric,li2022dense,zhang2022isnet}.
For SIRST3, we merge the three subsets and perform a 6:2:2 split for train/val/test to evaluate robustness on diverse scenes.

\noindent\textbf{Experimental settings.} We trained for 400 epochs using the AdamW optimizer with a batch size of 16 and a learning rate of $1 \times 10^{-3}$. Outer-level update activates at epoch 80, updating the sample-weighting and pseudo-label branches in parallel. The pseudo-label branch updates every 20 epochs via a separate Adam optimizer with learning rate $1 \times 10^{-2}$. Inputs were normalized and randomly cropped to $256 \times 256$. Soft IoU loss is used to define $\mathcal{L}_\mathtt{val}$ and $\mathcal{L}_\mathtt{tr}$.
Both the training and validation sets are supervised only by single-point annotations, and all image priors used in our method are computed solely from the input infrared images.
We report standard SIRST metrics (IoU, nIoU, P$_{d}$, F$_{a}$) to assess segmentation fidelity, missed targets, and false positives.

\subsection{Experimental Results}
\noindent\textbf{Evaluation on the SIRST3 dataset.}
Table~\ref{tab:major1} presents a comprehensive performance comparison of our proposed training strategy against leading single-point supervision baselines (LESPS, PAL, MCLC) and a fully supervised (Full) upper bound. The evaluation spans seven different detector architectures on the comprehensive SIRST3 benchmark. The results definitively demonstrate that our approach consistently and significantly outperforms all competing single-point supervision methods across nearly every metric and detector. This trend of superiority is consistent across all benchmarks; on SIRST3, our strategy achieves the highest IoU and nIoU for every evaluated detector among all non-full supervision methods.

Fig.~\ref{fig:radiation1} provides a detailed qualitative comparison of our method against typical single-point supervision baselines (LESPS, MCLC, and PAL) on challenging examples from the SIRST3 dataset. The analysis highlights a clear pattern of robustness for our approach. In high-clutter environments where targets are adjacent to bright edges or complex structures (Rows 1 and 2), competing methods struggle significantly. A multi-target case is shown in the third row, where our method separates targets cleanly and avoids merging them into surrounding structures, while preserving target boundaries more effectively overall. A low-contrast multi-target scenario is shown in the last row, where our method achieves higher detection accuracy by successfully identifying faint targets that other methods often miss.

\begin{figure}[t]
  \centering
  \includegraphics[width=\linewidth]{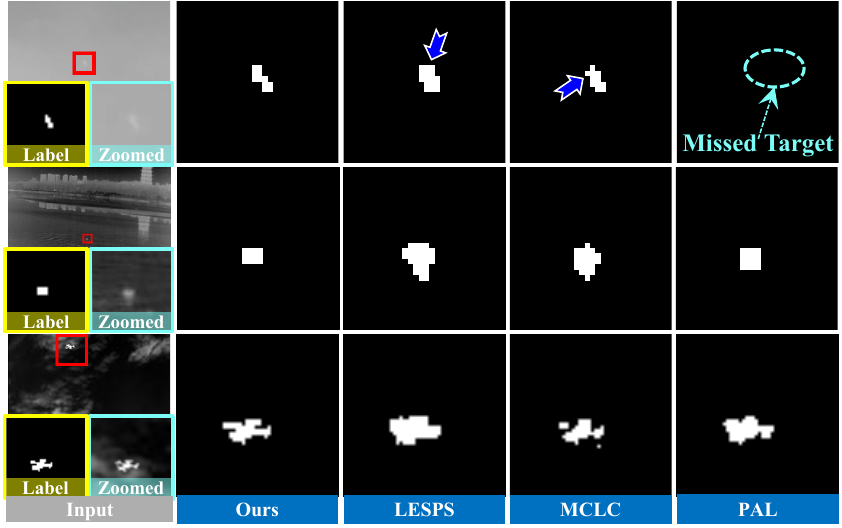}
  \caption{Qualitative comparison on three representative scenes. These results are obtained by ALCLNet.}
  \label{fig:radiation}
\end{figure}

\noindent\textbf{Evaluation on three individual datasets.}
We validated our framework's stability and robustness both qualitatively (Fig.~\ref{fig:radiation}) and quantitatively (Table~\ref{tab:major2}) on individual datasets. In challenging high-clutter and low-SNR scenes, our method reliably localizes targets while suppressing noise. In contrast, baselines like LESPS, MCLC, and PAL suffer from unstable evolution, missed detections, and false alarms. This stability is quantitatively confirmed on scarce-data benchmarks, where our method consistently and significantly outperforms all competitors; for example, achieving over 20 percentage points of IoU improvement over LESPS.

\subsection{Ablation Studies}
\noindent\textbf{Efficiency of the proposed framework.} 
Table~\ref{tab:wo_components} presents an ablation analysis of the core components of our framework on SIRST3. Two obvious conclusions can be obtained. First, supervision quality is the primary bottleneck: Combining diffusion with the superpixel prior improves all metrics, suggesting complementary effects of diffusion expanding supervision, while the superpixel prior regularizes boundaries and suppresses label noise. Second, the training strategy is essential for turning better pseudo-labels into stable optimization: removing the bi-level dual-update, balance, online label update, or dynamic aggregation (B1-B4) consistently degrades performance. Overall, these results support that effective pseudo-label construction and stable bi-level optimization are both necessary, and their combination leads to the best performance.

\begin{figure}[t]
  \centering
  \includegraphics[width=\linewidth]{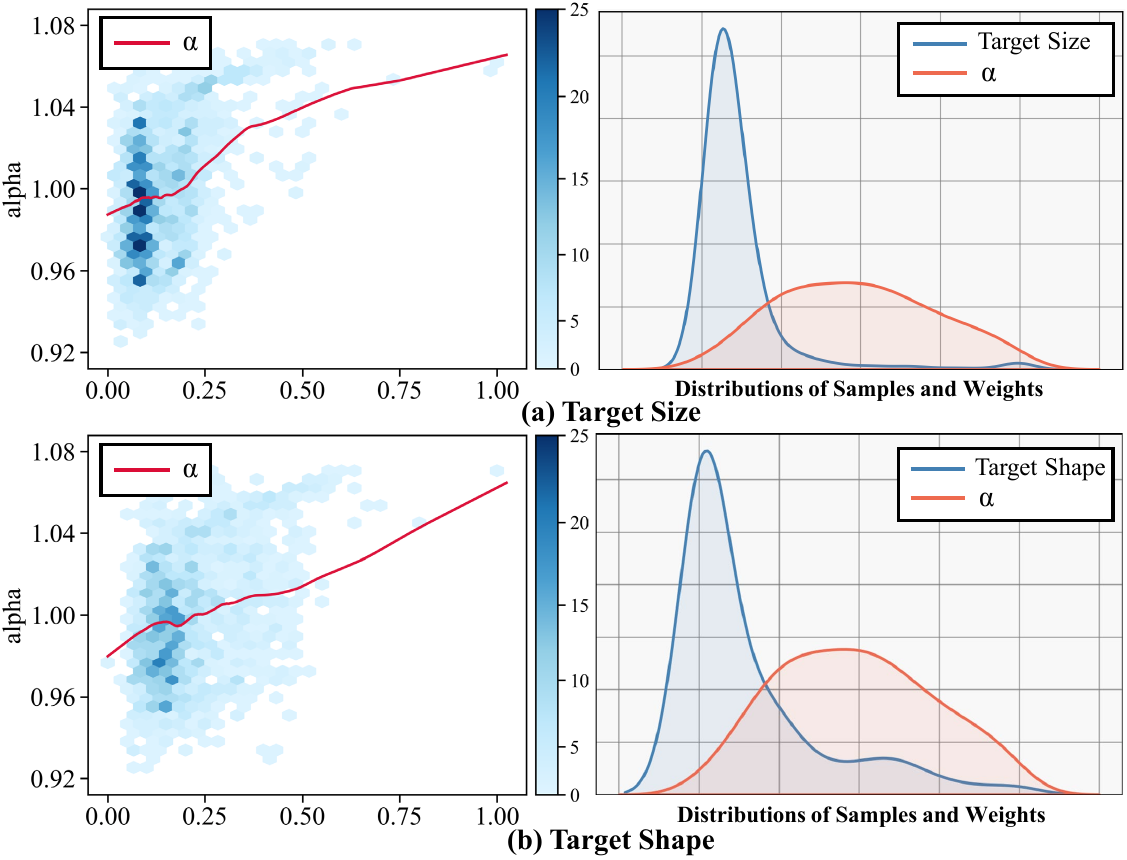}
  \caption{Analysis of the learned sample weight $\alpha$. The left plot shows the correlation of $\alpha$ with target size (small to large diameter) and target shape (near-circular to irregular), while the right plot shows the distribution of target shape, target size and $\alpha$.}
  \label{fig:alpha_distribution}
\end{figure}

\definecolor{CellBlue}{HTML}{E7EEF8}
\definecolor{CellGreen}{HTML}{FADADE}

\begin{table}[t]
    \centering
    \scriptsize
    \renewcommand{\arraystretch}{1.1}
    \setlength{\tabcolsep}{2.7mm}
    \caption{Ablation of core components in our framework. These results are obtained by ALCLNet.}
    \label{tab:wo_components}
    \begin{tabular}{|l|ccc|}
    \hline
    \multirow{2}{*}{Model Variants}
    & \multicolumn{3}{c|}{SIRST3} \\ \cline{2-4}
    & IoU$\uparrow$ & nIoU$\uparrow$ & P$_d$$\uparrow$ \\ \hline
    
    (A1) Naive point supervision & 8.07 & 17.51 & 78.02 \\
    (A2) Diffusion only & 69.64 & 70.80 & 91.82 \\
    (A3) Superpixel prior only & 68.64 & 69.41 & 93.16 \\
    (A4) Diffusion + Superpixel (numerical) & 70.75 & 71.47 & 94.50 \\ \hline
    
    (B1) Direct joint learning & 67.10 & 69.49 & 89.54 \\
    (B2) w/o balance ($\alpha\equiv1$) & 70.96 & 72.95 & 94.24 \\
    (B3) w/o online label update & 71.52 & 71.26 & 93.43 \\
    (B4) w/o dynamic aggregation ($\beta=0$) & 67.74 & 68.60 & 92.23 \\ \hline
    
    Proposed & \cellcolor{CellGreen}73.28 & \cellcolor{CellGreen}73.30 & \cellcolor{CellGreen}95.44 \\ \hline
    
    \hline
    \end{tabular}
\end{table}

\noindent\textbf{Verification of sample rebalancing.} 
To validate the meta-network's learned behavior, we conducted a statistical analysis of the weights $\alpha_i$ aggregated over 100 runs. As illustrated in Fig.~\ref{fig:alpha_distribution}, the weights correlate strongly with sample difficulty: challenging samples (\textit{e.g.,} larger, irregular targets) receive higher $\alpha_i$, while easy samples (\textit{e.g.,} small, compact targets) are assigned lower weights. Crucially, the weight density appears inversely correlated with the sample density; regions with high $\alpha$ density are precisely those where the sample density is low. The weighting branch automatically compensates for under-represented but challenging cases by upweighting them, while downweighting abundant cases.

\begin{figure}[t!]
    \centering
    \includegraphics[width=0.98\columnwidth]{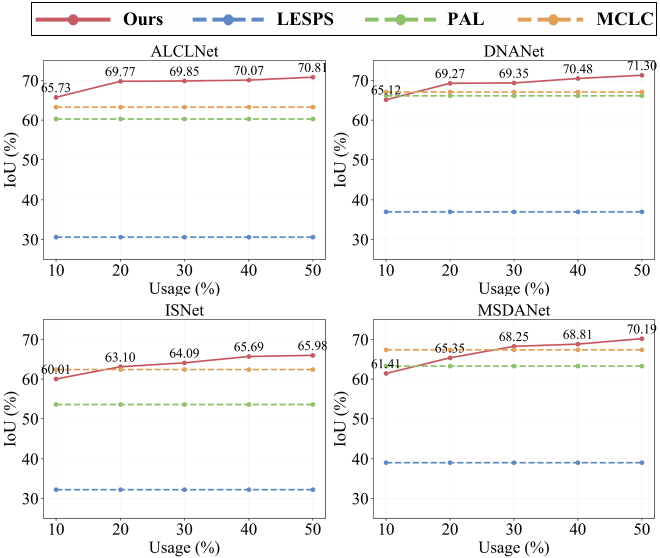} 
    \caption{Data-efficiency on SIRST3 dataset across diverse networks, comparing Ours, LESPS, PAL, and MCLC under single-point supervision. Others are trained with full training datasets.}
    \label{fig:datausage}
\end{figure}

\begin{figure}[t!]
    \centering
    \includegraphics[width=0.98\columnwidth]{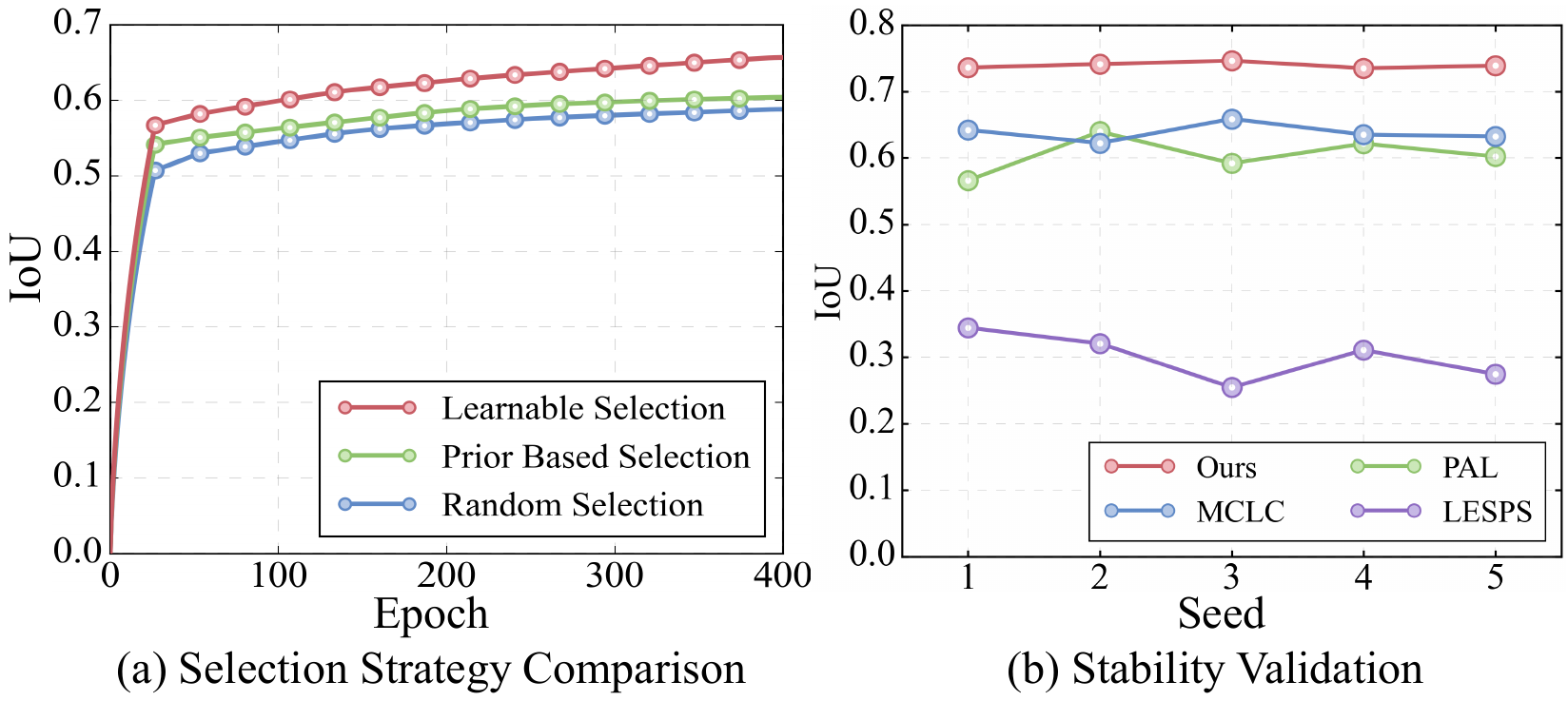} 
    \caption{Convergence of sample-selection strategy and stability validation on SIRST3 under diverse seeds.}
    \label{fig:selection}
\end{figure}

\noindent\textbf{Validation of data usage.} We found that the learned weights $\alpha_i$ can also be used to guide data selection, reducing annotation and training costs. Experiments demonstrate that a standard detector trained from scratch using only the top 30\% of the sample subset, selected via $\alpha_i$ ranking, already surpasses the performance of baselines trained on 100\% of the data, as shown in Fig.~\ref{fig:datausage}. This indicates that our bi-level optimization framework can effectively identify informative samples and safely discard redundant data.
Fig.~\ref{fig:selection} (a) compares three data selection strategies under the same data budget. Our learned weights clearly outperform random and prior-based (hand-crafted difficulty) selection, showing more reliable identification of informative samples. Fig.~\ref{fig:selection} (b) reports results under five random seeds. Our method exhibits smaller variance than the baselines, demonstrating significantly improved training stability.

\begin{figure}[t!]
    \centering
    \includegraphics[width=0.98\columnwidth]{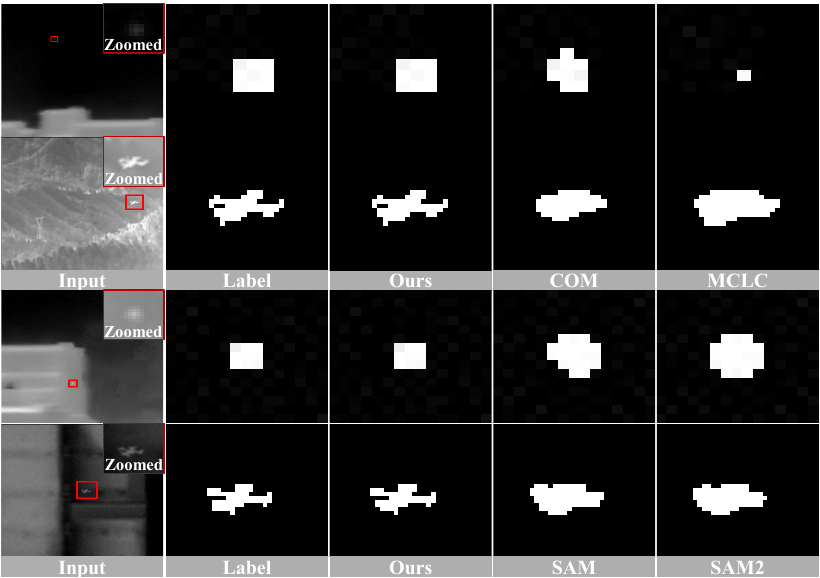} 
    \caption{Qualitative comparison of small target detection on the SIRST3 dataset: Ours vs. COM, MCLC, SAM, and SAM2.}
    \label{fig:com}
\end{figure}

\noindent\textbf{Time efficiency of  diffusion annotation.} 
We evaluated our proposed physics-induced diffusion annotation against existing single-point methods (COM and MCLC) in terms of both quality and efficiency.
Moreover, as shown in Table~\ref{tab:single_point}, our pseudo-masks achieve an IoU of up to 68.02 against ground truth on the SIRST3 dataset, outperforming MCLC (66.98) and COM (15.62). As shown in Fig.~\ref{fig:com}, visual comparisons also confirm that while COM tends to under-segment and MCLC often over-grows into the background, our method better maintains clean target boundaries.
In terms of efficiency, our method requires only 0.012 seconds per target on a CPU, which is approximately five times faster than MCLC (0.069s) and nearly two orders of magnitude faster than COM (0.923s).
This high efficiency (a five-fold speedup over MCLC) is crucial: it makes the diffusion module lightweight enough to be embedded within the bi-level optimization loop.
We compared our proposed physics-diffusion annotation with foundation models SAM/SAM2 in Table~\ref{tab:SAM}. Visually, SAM/SAM2 perform poorly on infrared small targets, often either missing them or causing severe over-segmentation, as shown in Fig.~\ref{fig:com}. In contrast, our method generates compact masks that better follow the target's true thermal footprint. In terms of efficiency, our diffusion module is also significantly faster than SAM/SAM2. This demonstrates that for single-point supervised ISTD, a dedicated physics-guided framework remains superior to generic foundation models in both accuracy and efficiency.

\definecolor{CellBlue}{HTML}{E7EEF8}
\definecolor{CellPink}{HTML}{FADADE}

\begin{table}[t!]
    \centering
    \renewcommand{\arraystretch}{1.1}
    \caption{Pseudo-mask generation on four ISTD datasets with Centroid and Coarse initialization, comparing COM, MCLC, and Ours, and reporting IoU and average time per target.}
    \label{tab:single_point}
    \resizebox{\linewidth}{!}{
    \begin{tabular}{|c|c|c|c|c|c|c|}
    \hline
    \multirow{2}{*}{Centroid} & SIRST3 & SIRST-v1 & NUDT-SIRST &IRSTD-1k & \multirow{2}{*}{Average Time↓} \\
    \cline{2-5}
     & IoU↑ & IoU↑ & IoU↑ & IoU↑ & \\
    \hline
    \multirow{1}{*}{COM~\cite{li2024levelset}} & 15.62 & 21.72 & 7.80 & 36.86 & 0.923s \\
    
    \multirow{1}{*}{MCLC~\cite{li2023monte}} & 66.98 & \cellcolor{CellPink}75.50 & 62.97 & 68.65 & 0.069s \\
    \cline{1-1}
    \multirow{1}{*}{Ours} & \cellcolor{CellPink}68.02 & {74.58} & \cellcolor{CellPink}64.48 & \cellcolor{CellPink}70.12 &\cellcolor{CellPink}0.012s \\
    \hline  
    \multirow{2}{*}{Coarse} & SIRST3 & SIRST-v1 & NUDT-SIRST & IRSTD-1k & \multirow{2}{*}{Average Time↓} \\
    \cline{2-5}
     & IoU↑ & IoU↑ & IoU↑ & IoU↑ & \\
    \hline
    \multirow{1}{*}{COM~\cite{li2024levelset}} & 15.01 & 20.76 & 7.67 & 35.11 & 0.961s \\
    
    \multirow{1}{*}{MCLC~\cite{li2023monte}} & 63.71 & \cellcolor{CellPink}74.19 & 58.47 & 66.18 & 0.069s \\
    \cline{1-1}
    \multirow{1}{*}{Ours} &\cellcolor{CellPink}65.69 & {73.78}&\cellcolor{CellPink}59.15 & \cellcolor{CellPink}68.42 & \cellcolor{CellPink}0.013s \\
    \hline
    \end{tabular}}
\end{table}

\begin{table}[t!]
    \centering
    \renewcommand{\arraystretch}{1.1}
    \caption{Pseudo-mask generation on four ISTD datasets with Centroid and Coarse initialization, comparing SAM, SAM2, and Ours, reporting IoU and average time per target.}
    \label{tab:SAM}
    \resizebox{\linewidth}{!}{
    \begin{tabular}{|c|c|c|c|c|c|c|}
    \hline
    \multirow{2}{*}{Centroid} & SIRST3 & SIRST-v1 & NUDT-SIRST & IRSTD-1k & \multirow{2}{*}{Average Time↓} \\
    \cline{2-5}
     & IoU↑ & IoU↑ & IoU↑ & IoU↑ & \\
    \hline

    \multirow{1}{*}{SAM~\cite{kirillov2023segany}} & 44.98 & 55.39 & 41.54 & 45.41 & 0.105s \\
    
    \multirow{1}{*}{SAM2~\cite{ravi2024sam2}} & 47.39 & 57.01 & 47.91 & 43.31 & 0.156s \\                 
    \cline{1-1}
     \multirow{1}{*}{Ours} &\cellcolor{CellPink}68.02 &\cellcolor{CellPink}74.58 &\cellcolor{CellPink}64.48 &\cellcolor{CellPink}70.12 &\cellcolor{CellPink}0.012s \\
     \hline  
    \multirow{2}{*}{Coarse} & SIRST3 & SIRST-v1 & NUDT-SIRST & IRSTD-1k & \multirow{2}{*}{Average Time↓} \\
    \cline{2-5}
     & IoU↑ & IoU↑ & IoU↑ & IoU↑ & \\
     \hline
    \multirow{1}{*}{SAM~\cite{kirillov2023segany}} & 6.03 & 4.41 & 7.98 & 4.13 & 0.135s \\
    
    \multirow{1}{*}{SAM2~\cite{ravi2024sam2}} & 47.14 & 57.06 & 47.67 & 42.95 & 0.149s \\
    \cline{1-1}
    \multirow{1}{*}{Ours} & \cellcolor{CellPink}65.69 &\cellcolor{CellPink}73.78 &\cellcolor{CellPink}59.15 &\cellcolor{CellPink}68.42
    &\cellcolor{CellPink}0.013s \\
    \hline
    \end{tabular}}
\end{table}
                
\section{Conclusion}
In this paper, we proposed a bi-level dual-update framework built upon single-point supervision to address annotation costs and sample imbalance jointly. First, we introduced a physics-induced annotation strategy. Second, we addressed the joint optimization of the detector, sample importance weights, and annotation quality. Extensive experiments demonstrated that our approach is highly efficient, achieving  superior detection accuracy.

\section*{Impact Statement}

This paper presents work whose goal is to advance the field of Machine
Learning. There are many potential societal consequences of our work, none
which we feel must be specifically highlighted here.

\section*{Acknowledgements}
This work is partially supported by the National Natural
Science Foundation of China (Nos.U22B2052,
624B2033), Central Guidance for Local Science and Technology
Development Fund (Youth Science Fund Project,
Category A, No. 2025JH6/101100001), the Distinguished
Young Scholars Funds of Dalian (No.2024RJ002), and the Fundamental
Research Funds for the Central Universities.

\bibliography{main}
\bibliographystyle{icml2026}

\end{document}